\documentclass[letterpaper, 10 pt, conference]{ieeeconf}  

\IEEEoverridecommandlockouts                              

\usepackage{graphics} 
\usepackage{epsfig} 
\usepackage{mathptmx} 
\usepackage{times} 
\usepackage{amsmath} 
\usepackage{amssymb}  
\usepackage{hyperref}
\usepackage{multirow}
\usepackage{multicol}
\usepackage{booktabs}
\usepackage{cite}

\usepackage[mathscr]{eucal}
\title{\LARGE \bf
Robust Navigation with Cross-Modal Fusion and Knowledge Transfer
}
\author{Wenzhe Cai$^{\dagger }$, Guangran Cheng$^{\dagger}$, Lingyue Kong, Lu Dong, Changyin Sun$^{*}$ 
\thanks{$\dagger$ Equal Contribution}
\thanks{* Corresponding Author}
\thanks{Wenzhe Cai, Guangran Cheng, Lingyue Kong and Changyin Sun are with the school of Automation, Southeast University, Nanjing, 210096, China,
        Lu Dong is with the school of Cyber science and Engineering, Southeast University, Nanjing, 210096, China, 
        Emails: \{wz\_cai,chenggr,lingyuekong,ldong90,cysun\}@seu.edu.cn.
        }%
\thanks{This paper is supported by the National Key Research and Development Program of China under Grant 2018AAA0101400, the National Nature Science Foundation of China under Grant 61821004, 62173251, and the Nature Science Foundation of Jiangsu Province of China under Grant BK20202006.}%
}
\begin{document}
\maketitle
\pagestyle{empty}  
\thispagestyle{empty} 
\thispagestyle{empty}
\pagestyle{empty}

\begin{abstract}
    Recently, learning-based approaches show promising results in navigation tasks.
    However, the poor generalization capability and the simulation-reality gap prevent a wide range of applications.
    We consider the problem of improving the generalization of mobile robots and achieving sim-to-real transfer for navigation skills.
    To that end, we propose a cross-modal fusion method and a knowledge transfer framework for better generalization.
    This is realized by a teacher-student distillation architecture.
    The teacher learns a discriminative representation and the near-perfect policy in an ideal environment.
    By imitating the behavior and representation of the teacher, the student is able to align the
    features from noisy multi-modal input and reduce the influence of variations on navigation policy.
    We evaluate our method in simulated and real-world environments.
    Experiments show that our method outperforms the baselines by a large margin and achieves robust navigation performance with varying working conditions.
\end{abstract}
\section{Introduction}
While the SLAM-based traditional navigation approaches\cite{LaValle2006PlanningA,Thrun2002ProbabilisticR,Wrobel2001MultipleVG} enable robot with navigation skills, they largely depend on complicated manually-designed modules. 
It may become fragile in environments with dynamic objects, pose estimation errors, and low texture.
The learning-based methods, with concise end-to-end training architecture and powerful deep neural networks, attract diverse types of studies in navigation problems.
Many representative works are proposed in point-to-point navigation\cite{Wijmans2020DDPPOLN,Zhao2021TheSE}, object-goal navigation\cite{Qiu2020LearningHR,Chaplot2020ObjectGN}, visual-language navigation\cite{Wang2021StructuredSM,chen2021history}.
Most works model the navigation problem as a Markov Decision Process (MDP) and refer to deep reinforcement learning (DRL) algorithms for decision-making.
Considering the high sampling costs in the trial-and-error learning process and low data efficiency in DRL, it is expensive to train the robot in real world.
However, the zero-shot transfer to reality can lead to unexpected or even dangerous consequences.
Therefore, improving the generalization of learning-based navigation methods is important for bridging the sim-to-real gap.

\begin{figure}
    \includegraphics[width=0.5\textwidth]{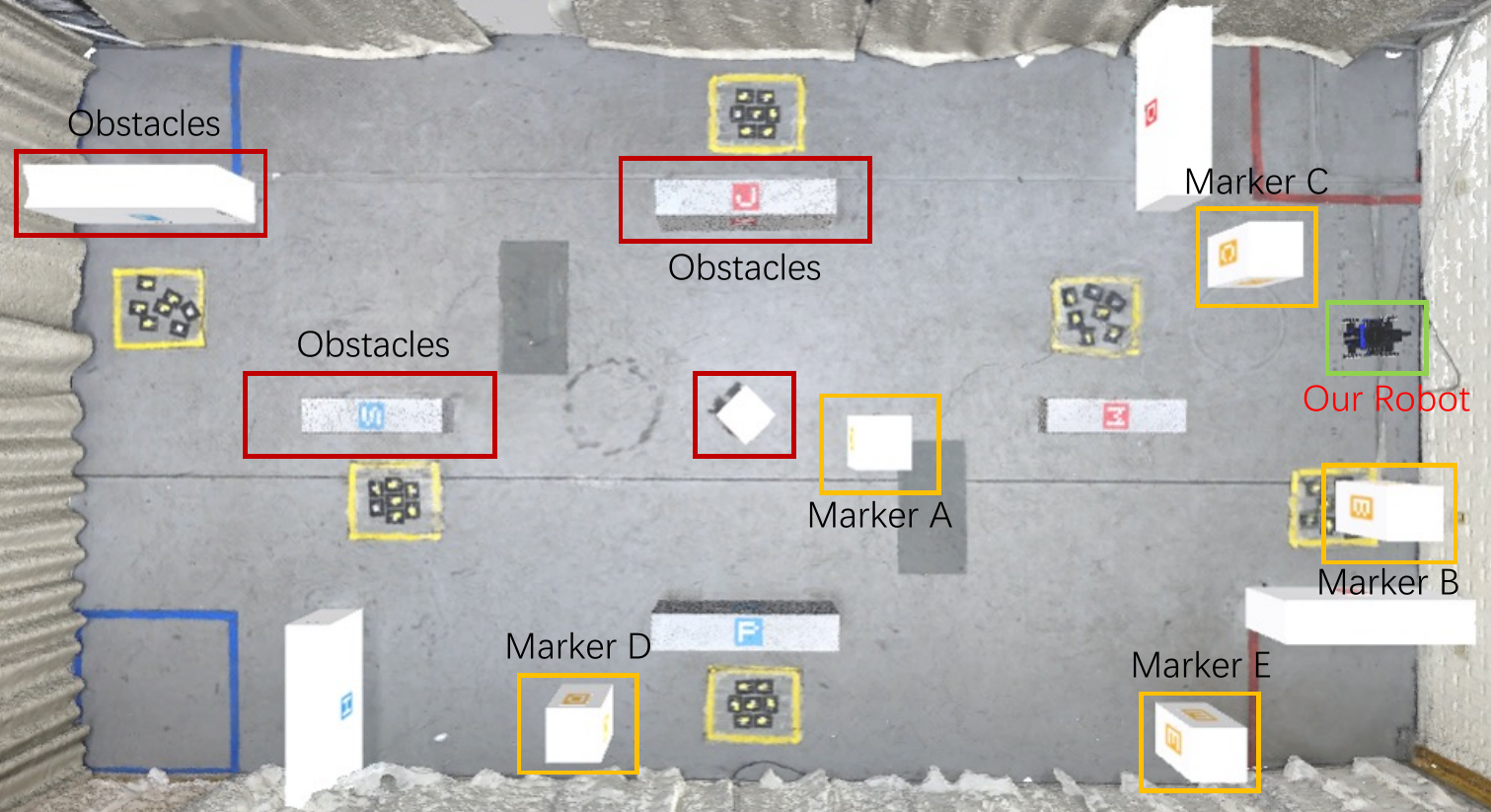}
    \caption{A brief introduction of the navigation task. 
             The robot (green rectangle) plans a navigation route to goals A,B,C,D,E in order.
             The initial position of of robot and goals (yellow rectangles) are randomized at each episode.
             Other obstacles are marked in red.}
    \label{FigureEnv}
    \vspace{-0.5cm}   
\end{figure}

In real-world navigation systems, robots are usually equipped with multiple sensors (lasers, cameras, IMUs).
Accurate perception is a prerequisite to train a robust navigation policy.
To deal with ubiquitous sensor noise, researchers point out that multi-modal observations provide complementary properties\cite{feng2020deep,Liang2018DeepCF}.
Therefore, it is possible to detect and identify data misalignment and error with advanced multi-sensor data fusion methods.
Another important class of methods skips the explicit denoising procedure but concentrates on improving the policy generalization.
Lots of deep learning-based approaches\cite{arndt2020meta, matas2018sim} use domain randomization to avoid overfitting the training set. 
This is proved to be an effective way to acquire more robust policies. 
However, domain randomization methods may corrupt the distributions of data in the original state space.
Since the exploration ability is a bottleneck for deep reinforcement learning algorithms, 
it adds the difficulty for DRL methods to discover the most task-relavant patterns, which may lead to sub-optimal policies.

\begin{figure*}[!ht]
    \centering
    \includegraphics[width=1.0\textwidth]{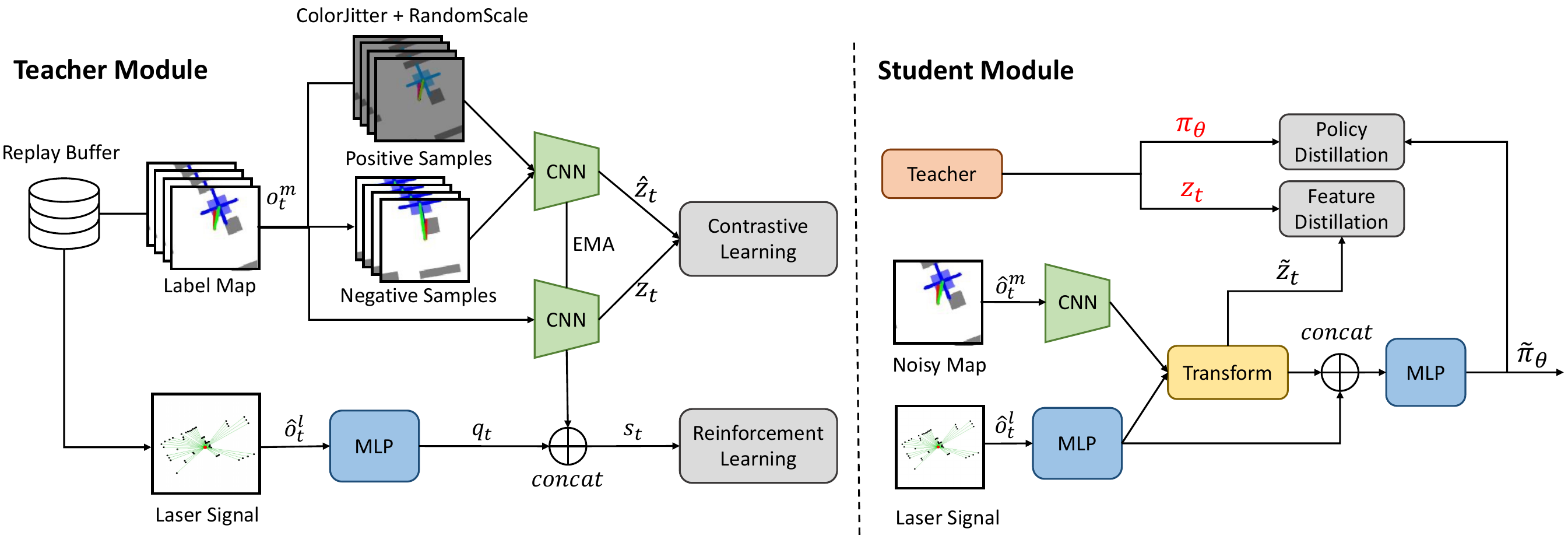}
    \caption{
    Teacher-Student Distillation Framework of our method.
    To train a teacher policy, we use label map observation $o^{m}_{t}$ and noisy laser observation $\hat{o}^{l}_{t}$ as policy inputs.
    In order to extract a useful representation that is sensitive to the agent pose information,
    we add a contrastive objective to widen the gap between two extracted features $\hat{z}_{t}$ and $z_{t}$ of the augmented images, which are different in relative position information.    
    Similar to MoCo\cite{He2020MomentumCF}, we use exponential moving average (EMA) to update one of the encoders.
    With the noisy input of map $\hat{o}^{m}_{t}$ and laser $\hat{o}^{l}_{t}$, the student is trained with supervised learning and tries to imitate the output from the encoder network $z_{t}$ and policy network $\pi_{\theta}$ of the teacher.
    Finally, the student is able to learn a transform function to align the feature $\widetilde{z}_{t}$ to $z_{t}$ and imitate the teacher behavior with $\widetilde{\pi}_{\theta}$.
    }
    \label{FigureArch} 
    \vspace{-0.5cm}  
\end{figure*}

To deal with sensor noises and learn a generic navigation policy, 
we propose a cross-modal fusion network that is capable of extracting both task-relevant and noise-sensitive features. 
We train a robust navigation policy with expert knowledge transfer. 
This is completed by a teacher-student distillation framework. 
Concretely, the teacher module is responsible for generating an expert navigation policy.
We train an RL agent in the ideal simulated environment (without sensor noise) as the expert policy without additional prior knowledge. The student module is designed to imitate the teacher's ways to perceive and react in the ideal environment to perform navigation with the existence of sensor noise.
By introducing such a two-phase training procedure, the student module is able to better understand the dynamics and environment transitions, then learns a more generalized navigation policy. 
In addition, the student module also learns a transform function to restore the original signals from multi-modal inputs to resist the noise. To encourage learning of discriminative representations with respect to sensor noise, we introduce an auxiliary contrastive objective for teacher module. 
As pointed out in \cite{Datta2020IntegratingEL}, self-localization accuracy is important for navigation.
In this work, we focus on the pose estimation error and discuss the generalization of our method in navigation scenarios as shown in Figure \ref{FigureEnv}. We evaluate our method both simulated and physical environments. The results show that our method improves the generalization with a large margin with varying working conditions (speed limit, noise scale, scene layout). 

In a nutshell, our contributions are summarized as follows:
\begin{itemize}
    \item We propose an expert knowledge transfer approach for robust robot navigation, which is completed with a teacher-student distillation framework.
    \item By introducing contrastive objective and feature distillation, our proposed cross-modal fusion approach can learn noise-sensitive representation and compensate for sensor noise.
    \item Empirical studies show the robustness of our navigation policy across various working conditions.
\end{itemize}

\section{Related Work}
\subsection{DRL-based Navigation}
Recently, many works employ DRL for navigation tasks.
Some methods follow the traditional SLAM frameworks but replace crucial components (e.g. localization, map building, path-planning modules) with deep learning methods.
This helps realize long-distance navigation in complex environments \cite{faust2018prm,francis2020long}.
Other methods collect data in simulators and directly learn monolithic navigation policies without planning guidance \cite{chiang2019learning,shi2019end}.
To improve the data efficiency, task-relevant auxiliary objectives are introduced in addition to the origin RL objective \cite{mirowski2016learning,Jaderberg2017ReinforcementLW,kulhanek2019vision}.
Similar to the prior works, we also model navigation as an RL problem, 
but we use a two-phase training framework that contains both reinforcement learning and supervised learning.
This outperforms the agent trained only with reinforcement learning.

\subsection{Multi-Modal Fusion}
In realistic navigation problems, an important issue is how to represent and fuse the multi-sensor observations in an appropriate way.
Some methods stack the feature maps along the depth or concatenate them as flattened vectors before they are advanced to the outputs of the downstream task \cite{chadwick2019distant,dou2019seg}. 
To explore the relative informativeness of different sensing modalities, Mixture of Experts (MoE) approaches process each feature map by domain-specific networks and model their weights explicitly \cite{valada2020self}. 
We propose a multi-modal fusion module to compensate for the noise. 
Different from the prior works, our fusion module explicitly learns a transform function and restored the original features.

\subsection{Generalization of Deep Reinforcement Learning}
Domain randomization is an important technique to improve the generalization of DRL methods.
With randomized dynamical attributes in simulation, the agent can learn the policy in a diverse set of samples, which this helps learn a more robust policy \cite{anderson2021sim,tobin2017domain,choi2019deep}.
Data augmentation can be regarded as a specific type of domain randomization method and many works refer to it for improving generalization\cite{Hansen2021GeneralizationIR, Kostrikov2021ImageAI}.
For example, by replacing the background texture, SODA\cite{Hansen2021GeneralizationIR} increases the success rate on pixel-to-control tasks.
Policy distillation is also an effective way to improve generalization by extracting knowledge from an experienced expert\cite{traore2019continual, zhou2020domain}.
Our work is inspired by policy distillation methods. 
But we introduce a contrastive loss and this is proved to be necessary for better navigation policy.

\section{Approach}
\subsection{Problem Formulation}
We formulate the robot navigation problem as a Markov Decision Process (MDP) defined by a tuple $(S,A,R,P,\gamma)$, where $S,A,P,R$ represents the state space, the action space, the reward function and, the transition function respectively.
$\gamma$ is a scalar and represents the discount factor. 
An optimal policy $\pi^{*}$ aims to maximum the discounted culmulative reward $G=E_{s\sim \pi}[\sum_{t=0}^{T}{\gamma^{t}R(s_{t},a_{t})}]$, where $a_{t}$ is sampled w.r.t the policy $\pi(s_{t})$.
Generally, instead of making decisions by the underlying state $s_{t}$, the agent can only access the sensor observations in navigation problems. Thus, 
it becomes a Partially Observable Markov Decision Process (POMDP). 
Concretely, we consider two types of observations for the navigation task, which are laser signal $o^{l} \in \mathbb{R}^{n}$ and egocentric occupancy map $o^{m} \in \mathbb{R}^{c\times h \times w}$.
Here we focus on how to perform robust navigation skills, therefore, we assume the global occupancy map is already constructed with existing methods. 
At each time step $t$, the agent estimates its own pose $\hat{p}_{t}=(\hat{x}_{t},\hat{y}_{t},\hat{r}_{t})$, where $\hat{x}_{t}$ and $\hat{y}_{t}$ represent the corresponding coordinates, and $\hat{r}_{t}$ represents the yaw angle.
The egocentric occupancy map is a cropped global map with the center at $\hat{p}_{t}$. 
Note that the noise in pose estimation error makes the occupancy map $o^{m}$ mistakenly reflect the surroundings, which tests the perceiving ability of the agent.
The navigation targets are indicated by five goal coordinates denoted as $g_{A,B,C,D,E} = \{(x_{A},y_{A}),...,(x_{E},y_{E})\}$.
By taking in the information from the noisy map, goal coordinates, and laser signals, the navigation policy needs to learn a mapping from observations to the robot control command $a_{t} \in \mathbb{R}^{d}$.

\subsection{Cross-Modal Fusion with Distillation}
Instead of training a navigation policy end-to-end with reinforcement learning, 
we propose a teacher-student distillation architecture for cross-modal fusion and robust policy learning.
This is achieved by an teacher reinforcement learning agent and a supervised student agent.
The illustration of our approach can be referred in Fig. \ref{FigureArch}.

\noindent \textbf{Teacher Module}:  
The teacher module learns navigation skills in environments without noise. 
Under such circumstances, the teacher is able to learn a near-perfect navigation policy $\pi_{\theta}$ under the deep reinforcement learning framework.
To train that policy, we use a label egocentric map $o^{m}_{t}$ without pose estimation noise, together with the laser signal $\hat{o}^{l}_{t}$ as the input observation space.
The egocentric map is processed by a 3 layers of convolutional neural networks (CNN) and the laser signal is processed by 2 layers of fully-connected network (FC).
Denote the feature from egocentric map as $z_{t}$ and the feature from laser as $q_{t}$.
The concatenated vector $(z_{t},q_{t})$ are regarded as the state $s_{t}$ for the actor-critic networks.
We train the teacher module with \textit{PPO}\cite{Schulman2017ProximalPO}.
To faciliate efficient training, we design a dense reward signal defined as follows:
\begin{equation}
    r_{t} = r^{step}_{t} + r^{col}_{t} + r^{goal}_{t} + \Delta d^{p}_{t}
\end{equation}
where $r^{step}_{t}=-0.01$ is a constant penalty, $r^{col}=-0.05$ is collision penalty, 
$r^{goal}=4.0$ when robot reaches a marker,  $\Delta d^{p}_{t}$ is the change of Eculidean distance to goal.

The DRL agent itself guarantees no improvement of generalization, especially for an agent trained with ideal conditions.
Therefore, we introduce a contrastive objective to learn a discriminative feature representation that is sensitive to pose estimation error.
We construct positive and negative pairs for egocentric maps and use \textit{InfoNCE}\cite{Oord2018RepresentationLW} loss for optimization.
Two augmented images from the same egocentric map $o^{m}_{t}$ are regarded as positive pairs $(q,k_{+})$ while the rest in the mini-batch compose the negative pairs $(q,k_{-})$.
In order to train a representation that can capture the pose information, we use \textit{ColorJitter} and \textit{RandomScale} as the data augmentation methods.
Thus the network will learn to neglect the variations in size and color but focus on the relative pose information on the map.
Similar to \textit{MoCo}\cite{He2020MomentumCF}, we use a momentum encoder to avoid mode collapse.
The auxiliary objective is proved to be crucial and a detailed ablation study are described in the experiment part.
The \textit{InfoNCE} loss is defined as:
\begin{equation}
    \mathcal{L}^{InfoNCE} = -\mathbb{E}[log\frac{exp(q\cdot k_{+}/\tau)}{\sum_{j=0}^{B}exp(q \cdot k_{-}/\tau)}]
\end{equation}
$\tau$ is a temperature hyper-parameter and we set $\tau=0.25$ here.
We balance the importance of the auxiliary objective with a coefficient $\beta$ and the final loss function of the teacher module 
is defined as follows:
\begin{equation}
    \mathcal{L}^{teacher} = \mathcal{L}^{PPO} + \beta^{t}\cdot \mathcal{L}^{InfoNCE}
\end{equation}
We set $\beta^{t}=0.2$ in our experiments. And $\mathcal{L}^{PPO}$ is the prposed reinforcement learning objective in \cite{Schulman2017ProximalPO}.
we run 4 parallel environments with 2M steps to train the teacher module.

\noindent \textbf{Student Module}: 
The student module contains a multi-modal fusion network and a policy network which are both trained with supervised learning.
We expect the student to imitate the teacher policy as similarly as possible.
To that end, we introduce a feature distillation objective as well as a policy distillation objective.
Specifically, the multi-modal fusion network takes in the noisy egocentric map $\hat{o}^{m}_{t}$ and the laser signal $\hat{o}^{l}_{t}$.
Information from the map $\hat{o}^{m}_{t}$ is processed with 3-layers of CNN and the laser signal is processed with 2-layers of FC.
Denote the feature of two modals as $\hat{z}^{t}$ and $\hat{q}^{t}$.
Since the teacher module has been able to extract task-relevant features, 
we train the student network to learn a multi-modal fusion function to predict the teacher feature $z_{t}$.
The fusion is completed by a transform module which contains 2 layers of FC.
For simplicity, we use FC layers to fuse the information and align with the teacher feature, but any advanced architecture can be equipped here.
As we train the agents in simulators, it is feasible to collect the paired observation data, one with noise and the other not.
Therefore, at each time step, we can calculate the discrepancy between the teacher's representation and the student representation.
Denote the transform layer as $T_{\theta}$, the feature distillation objective is defined as negative cosine similarity:
\begin{equation}
    \mathcal{L}^{FD} = -\mathbb{E}[\frac{z_{t}\cdot T_{\theta}(\hat{z}^{t},\hat{q}^{t})}{||z_{t}||\cdot ||T_{\theta}(\hat{z}^{t},\hat{q}^{t})||}] 
\end{equation}
Although we train a feature transformation function, tiny difference between the teacher representation and the student representation can be magnified by policy network.
As a result, we also use a policy distillation objective which is defined as follows:
\begin{equation}
    \mathcal{L}^{PD} = \mathbb{E}[KL(\pi^{t}_{\theta}(s_{t})||\pi^{s}_{\theta}(s_{t}))]
\end{equation}
We optimize a weighted loss function which is defined as:
\begin{equation}
    \mathcal{L}^{student} = \alpha^{s} \cdot \mathcal{F}^{FD} + \beta^{s} \cdot \mathcal{F}^{PD}
\end{equation}
We use $\alpha^{s}=0.25$ and $\beta^{s}=0.75$ in our experiments.
A detailed table of hyper-parameters are listed in Table \ref{TabHyp}.

\begin{table}
    \centering
    \caption{Hyper-parameters details of our method}
    \begin{tabular}{|c|c|c|}
         \hline
         &Teacher Module & Student Module \\
         \hline
         Optimization & Reinforcement Learning & Supervised Learning \\
         \hline
         Algorithm & PPO & - \\
         \hline
         Parallels & 4 & 2 \\
         \hline
         Training Steps & 2M & 1M \\
         \hline
         Optimizer & Adam & Adam \\
         \hline
         Discount Factor & 0.99 & - \\
         \hline
         GAE-$\lambda$ & 0.95 & - \\
         \hline
         PPO-CLIP & 0.15 & - \\
         \hline
         nsteps & 256 & 256 \\
         \hline
         nepochs & 2 & 2 \\
         \hline
         nminibatch & 4 & 2 \\
         \hline
         learning\_rate & 4e-4 & 2e-4 \\
         \hline
         lr\_rate\_decay & Linear & Linear \\
         \hline
    \end{tabular}
    \label{TabHyp}
    \vspace{-0.5cm}
\end{table}

\section{Experiment}
\subsection{Experiment Setup}
\begin{figure}
    \centering
    \includegraphics[width=0.48\textwidth]{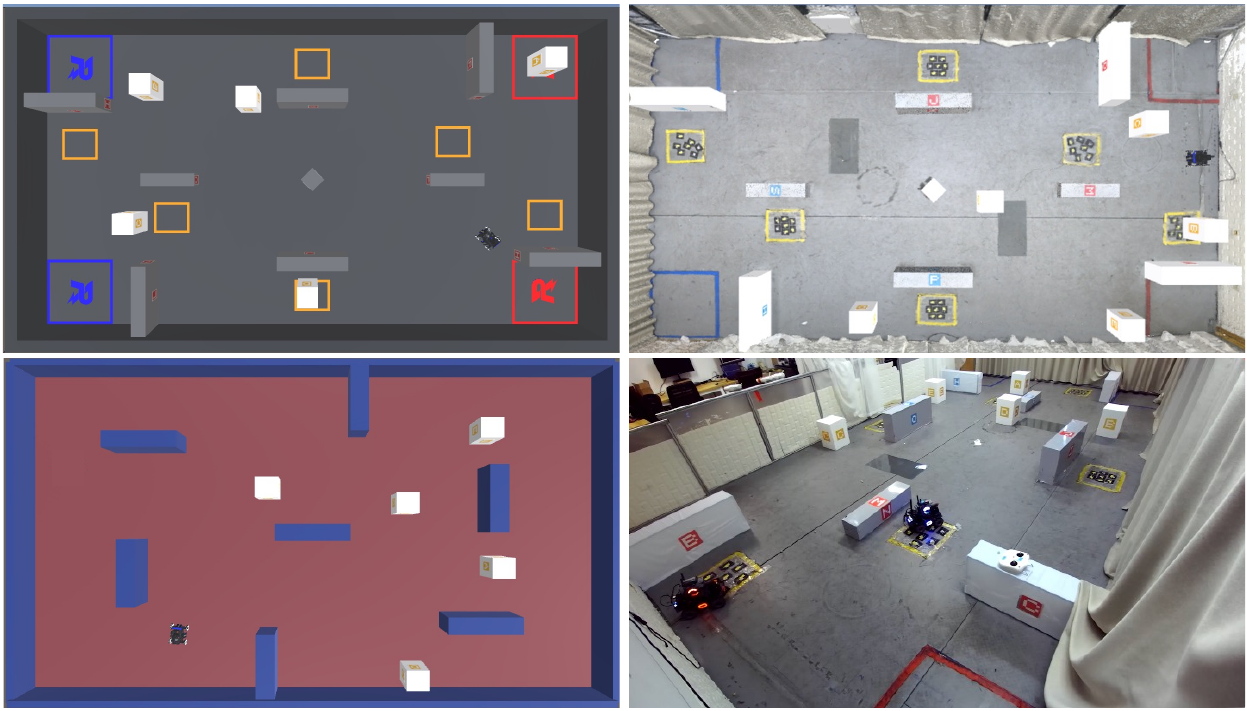}
    \caption{Training and Testing environments in our experiments.
             All the training is completed in the top-left environments.
             We test the generalization across speed limit and noise in the second top-right environment.
             The left-bottom environment is used for testing the generalization across scenes.
             The last is an overview of our robot navigation environment in the real world.}
    \label{ExpEnvs}
    \vspace{-0.5cm}
\end{figure}

The training environment is at the size of $8.08m \times 4.48m$.
The laser signal $o^{l}_{t}$ covers the detection range from $-135$ degree to $135$ degree with $N=60$ rays.
The egocentric map $o^{m}_{t}$ reflects the robot surroundings in occupancy map at the pose $\hat{p}_{t}$ in $2.56m \times 2.56m$.
The control command $a_{t}=(v^{x},v^{y},\omega)$ are 3-dim continuous vectors including the linear speed and angular speed along the robot Cartesian coordinate system.
To evaluate the generalization of our proposed method, we consider different variations between training and testing, including speed limit, noise scale and scene layouts.
Specifically, during the training, the maximum of speed is $(2m/s,2m/s,\frac{\pi}{4} rad/s)$.
Without loss of generality, we consider an episodic shift $\delta^{shift} \sim U(-0.5,0.5) $ and uniform noise $\delta^{p}_{t} \sim U(-0.1,0.1) $ as pose estimation error.
The laser signal is also injected with uniform noise $\delta^{l}_{t} \sim U(-0.1,0.1)$. 
$\delta^{shift}$ stays fixed during one episode but different across episodes.
The uniform noise $\delta^{p}$,$\delta^{l}$ is variant at each time step.
Five navigation goals are represented by $(x,y)$ coordinates and the position of goals are randomized at each episode.
During the training, the robot is requested to navigate to goals following the alphabet order in 500 steps, roughly equal to 20 seconds of clock time.
The testing time is set to 60 seconds. 
It is worth to mention that the control mode between training and testing is different:
In training, the interaction between robot and environment obeys MDP setting, 
but in testing, all the interaction are happened real-time, not only the control command but the control frequency also influences the performance.
This result in slightly different state transition functions between training and testing.
An overview of environment structures in training and testing are shown in Fig \ref{ExpEnvs}.

\subsection{Evaluation Metrics}
We consider the following metrics to evaluate the generalization of different navigation methods:
\begin{itemize}
    \item \textbf{Success Rate:} The average count of episodes where the robot successfully navigates to the 5 goals in order within 60 seconds.
    \item \textbf{Activation Count:} The average count of the activated goals.
    \item \textbf{Navigation Time:} The average clock time used for the navigation task.
    \item \textbf{Collision:} The average clock time of collision in an episode. 
\end{itemize}

\subsection{Baselines Methods}
We implement three different approaches as the baselines, denoted as \textit{Pose+Laser}, \textit{Map+Laser}, \textit{Teacher}.
The details of the baselines are described as follows:
\begin{itemize}
    \item \textbf{Pose+Laser}: An RL agent that concatenates the features extracted from pose estimation and noisy laser signal as states for policy inputs. 
    \item \textbf{Map+Laser}: Different from the first method, it integrates the surrounding information and the robot pose estimation into an egocentric map.
                              And it uses the extracted feature from CNN to replace the pose feature for policy inputs.
    \item \textbf{Teacher}:  Different from the former two methods, it learns the navigation policy in an ideal environment without sensor noisy.
                             We directly test the zero-shot performance in the noisy environment.
\end{itemize}

\subsection{Generalization Across Speed Limit}
In this experiment, we consider two speed limit settings $(1m/s,1m/s,\frac{\pi}{4} rad/s)$, $(0.5m/s,0.5m/s,\frac{\pi}{6} rad/s)$.
We use the same noise settings in training.
Changing the speed limit raises the problem of dynamics mismatch and results in a new MDP which owns a different transition function.
Although two MDPs share common task-relevant attributes, the robot will encounter novel states that have not appeared.
We list the performance with two settings in Table \ref{TabSpeed}.
Only the \textit{Map + Laser} achieves the compatible performance in success rate and activation count, 
but it fails in collision avoidance, while our method outperforms all the baselines with a large margin in all the metrics.

\begin{table}
    \centering
    \caption{Performance on speed generalization.}
    \begin{tabular}{c|cccc}
        \toprule[1pt]
        \multicolumn{5}{c}{$v_{x}=1m/s,v_{y}=1m/s, \omega = \pi / 4 \ rad/s$} \\
        \midrule[0.5pt]
        \textbf{Methods} & Success(\%) & Activation & Collision(s) & NavTime(s)\\
        \midrule[0.5pt]
        Pose + Laser & 61.8 & 3.54 & 10.14 & 37.65 \\
        \midrule[0.5pt]
        Map + Laser & 90.6 & 4.72 & 8.37 & 26.48 \\
        \midrule[0.5pt]
        Teacher & 74.1 & 4.34 & 5.73 & 35.06 \\
        \midrule[0.5pt]
        Ours & \textbf{98.1} & \textbf{4.96} & \textbf{2.01} & \textbf{21.73} \\ 
    \end{tabular}
    \begin{tabular}{c|cccc}
        \toprule[1pt]
        \multicolumn{5}{c}{$v_{x}=0.5m/s,v_{y}=0.5m/s, \omega = \pi / 6 \ rad/s$} \\
        \midrule[0.5pt]
        \textbf{Methods} & Success(\%) & Activation & Collision(s) & NavTime(s)\\
        \midrule[0.5pt]
        Pose + Laser & 28.2 & 2.46 & 8.45 & 54.81 \\
        \midrule[0.5pt]
        Map + Laser & 84.1 & 4.46 & 10.51 & 45.18 \\
        \midrule[0.5pt]
        Teacher & 42.6 &  3.31 & 4.27 & 50.43 \\
        \midrule[0.5pt]
        Ours & \textbf{88.1} & \textbf{4.53} & \textbf{2.34} & \textbf{43.83} \\
        \bottomrule[1pt]
    \end{tabular}
    \vspace{-0.5cm}
    \label{TabSpeed}
\end{table}

Note that the cross-modal fusion module is designed to compensate for the sensor noise, 
but also shows better generalization ability with respect to the speed limit. 
It implies that policy distillation is beneficial to learn a better representation for environment dynamics.

\subsection{Generalization Across Noise}
\begin{figure}
    \centering
    \includegraphics[width=0.22\textwidth]{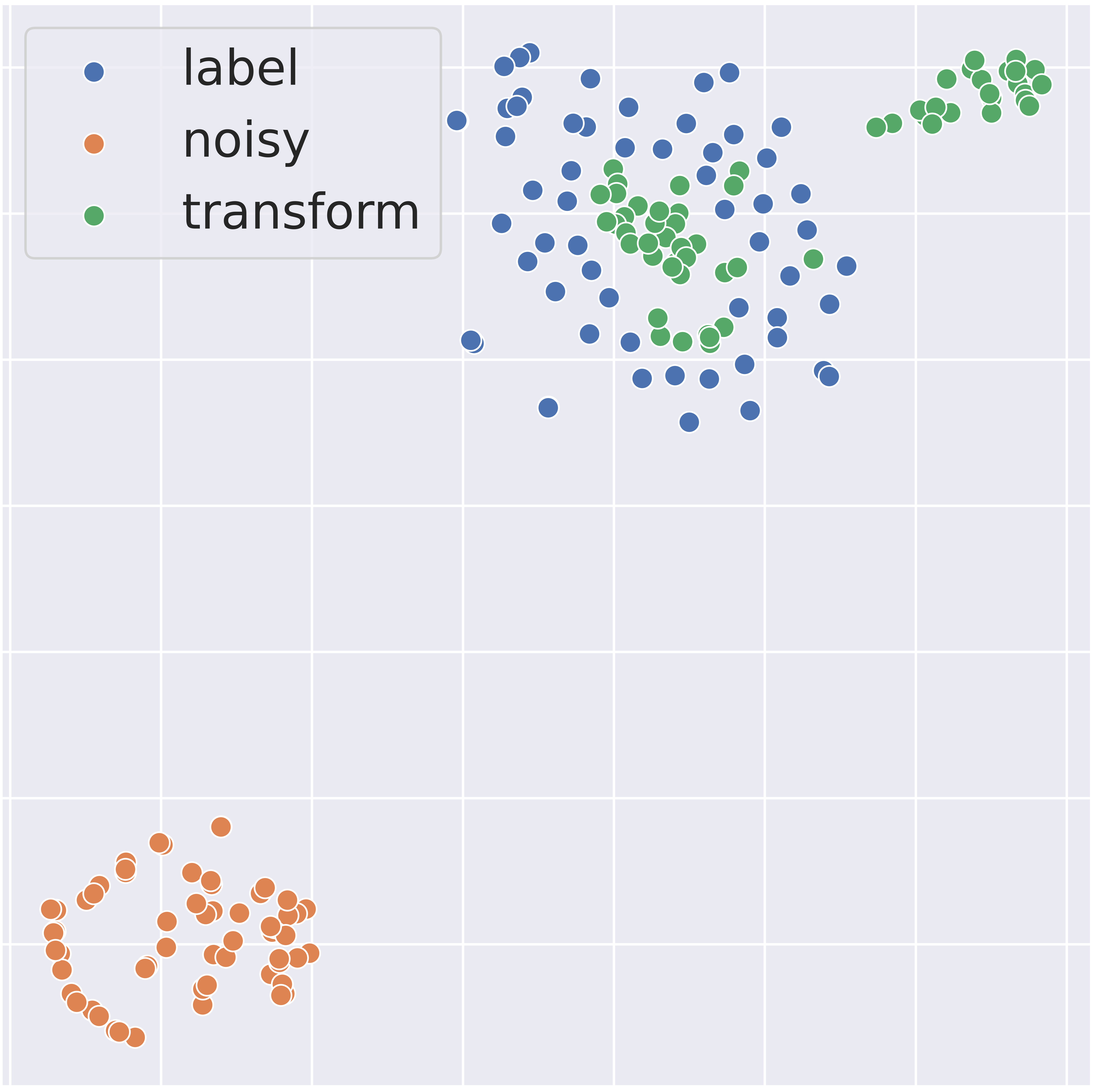}
    \includegraphics[width=0.22\textwidth]{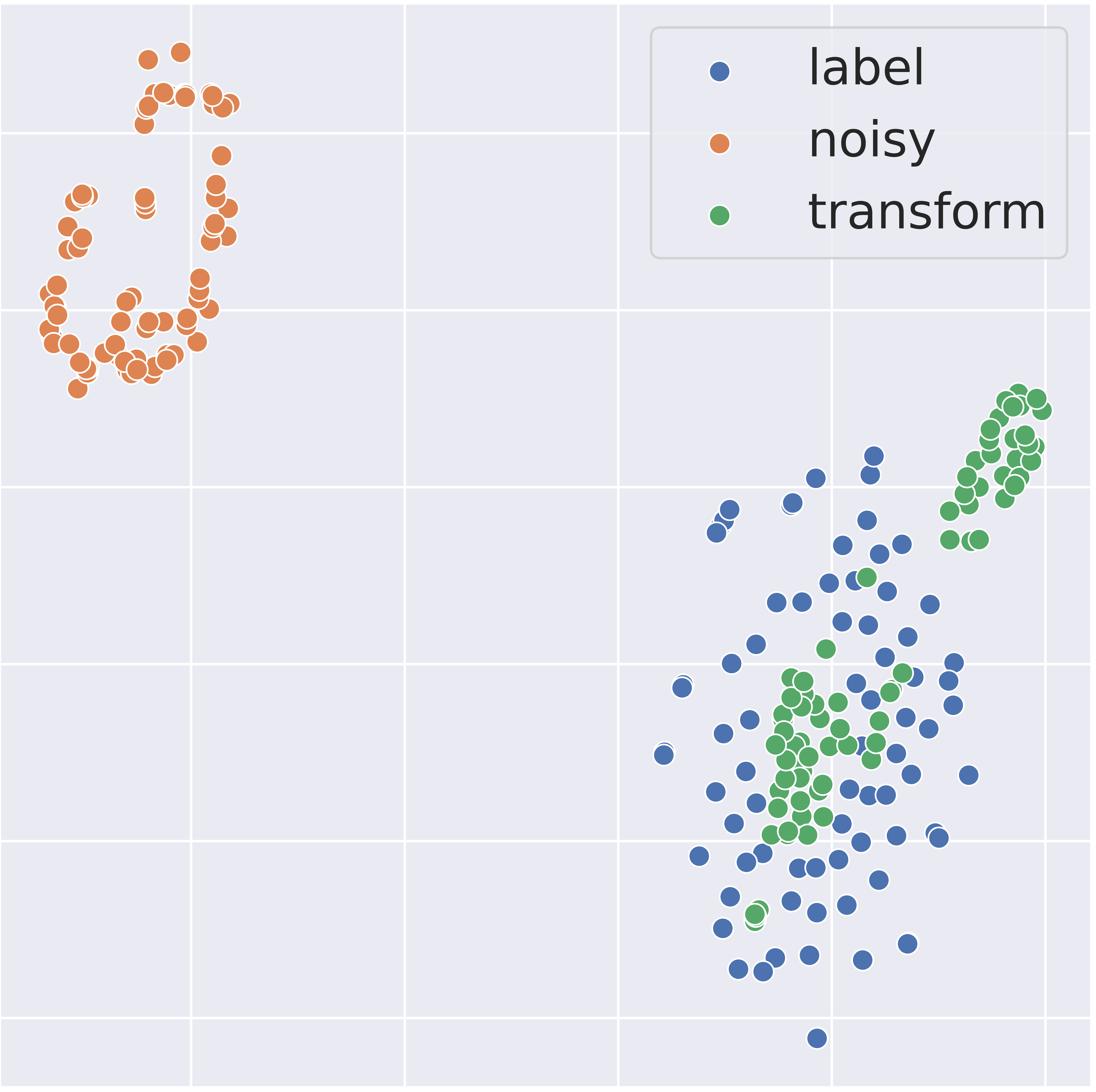}
    \caption{T-sne visualization of the embedding feature from the Label Map, the Noisy Map, and the Transformed Feature with Multi-Modal Fusion, which are represented by blue,orange,and green dots respectively. Our proposed multi-modal fusion method greatly reduces the gap to the label map.}
    \label{FigureTSNE}
    \vspace{-0.5cm}
\end{figure}

In this experiment, we fix the speed setting as $(1m/s,1m/s,\frac{\pi}{4}rad/s)$ and discuss the influence of the noise with different scales. The results are reported in Table \ref{TabNoise}.
Two different noise settings are considered here: 
In the easy scenarios, we set the episode shift $\delta^{shift} \sim U(-0.25,0.25)$ and uniform noise for pose estimation error as $\delta^{p}_{t} \sim U(-0.05,0.05)$.
The uniform noise for laser signal is also set to $\delta^{p}_{t} \sim U(-0.05,0.05)$. 
Most approaches achieve good performance (90+\% success rate) under this settings. 
But when we increase the noise, i.e. $\delta^{shift} \sim U(-0.75,0.75),\delta^{p}_{t} \sim U(-0.15,0.15),\delta^{l}_{t} \sim U(-0.15,0.15) $,
the performance of baseline methods degrade dramatically, especially for the teacher module (nearly 60\% drop on success rate).
Without training in noisy environments, the teacher policy cannot handle situations with unseen states.
Our method still maintains a satisfying performance, outperforming the best baselines \textit{Map+Laser} with 16.5\% on success rate and nearly a half collision time.
By imitating the teacher representation and policy, we reckon that it learns the transform function to restore the original feature by incorporating the information from two modals.
To verify our hypothesis, we visualize the features before and after the transform layers in the student module and compare it with the label feature extracted from the teacher module.
T-sne visualization of 3 types of features are shown in Fig \ref{FigureTSNE}.
Before feeding into the transform layer, the noisy feature (orange dots) is far from the label features (blue dots).
And the aligned feature (green dots) is close to the label features.
This shows that our cross-modal fusion approach can transform the noisy feature into the labels' neighborhood and reduce the influence of sensor noise.
\begin{table}
    \caption{Performance on noise generalization.}
    \centering
    \begin{tabular}{c|cccc}
        \toprule[1pt]
        \multicolumn{5}{c}{$\delta^{shift} \sim U(-0.25,0.25),\delta^{p}_{t} \sim U(-0.05,0.05),\delta^{l}_{t} \sim U(-0.05,0.05) $} \\
        \midrule[0.5pt]
        \textbf{Methods} & Success(\%) & Activation & Collision(s) & NavTime(s)\\
        \midrule[0.5pt]
        Pose + Laser & 71.6 & 4.19 & 10.96 & 35.01  \\
        \midrule[0.5pt]
        Map + Laser & 95.7 & 4.82 & 3.75 & \textbf{21.75}  \\
        \midrule[0.5pt]
        Teacher & 94.3 & 4.84 & 4.48 & 24.91  \\
        \midrule[0.5pt]
        Ours & \textbf{97.3} & \textbf{4.90} & \textbf{2.44} & 22.31 \\ 
    \end{tabular}
    \begin{tabular}{c|cccc}
        \toprule[1pt]
        \multicolumn{5}{c}{$\delta^{shift} \sim U(-0.75,0.75),\delta^{p}_{t} \sim U(-0.15,0.15),\delta^{l}_{t} \sim U(-0.15,0.15) $} \\
        \midrule[0.5pt]
        \textbf{Methods} & Success(\%) & Activation & Collision(s) & NavTime(s)\\
        \midrule[0.5pt]
        Pose + Laser & 34.2 & 2.84 & 13.34 & 48.57 \\
        \midrule[0.5pt]
        Map + Laser & 64.7 & 3.86 & 7.26 & 37.61  \\
        \midrule[0.5pt]
        Teacher & 34.1 & 2.64 & 4.16 & 49.87 \\
        \midrule[0.5pt]
        Ours & \textbf{81.2} & \textbf{4.59} & \textbf{3.45} & \textbf{32.48} \\
        \bottomrule[1pt]
    \end{tabular}
    \label{TabNoise}
\end{table}

\begin{figure*}[!ht]
    \centering
    \includegraphics[width=0.95\textwidth]{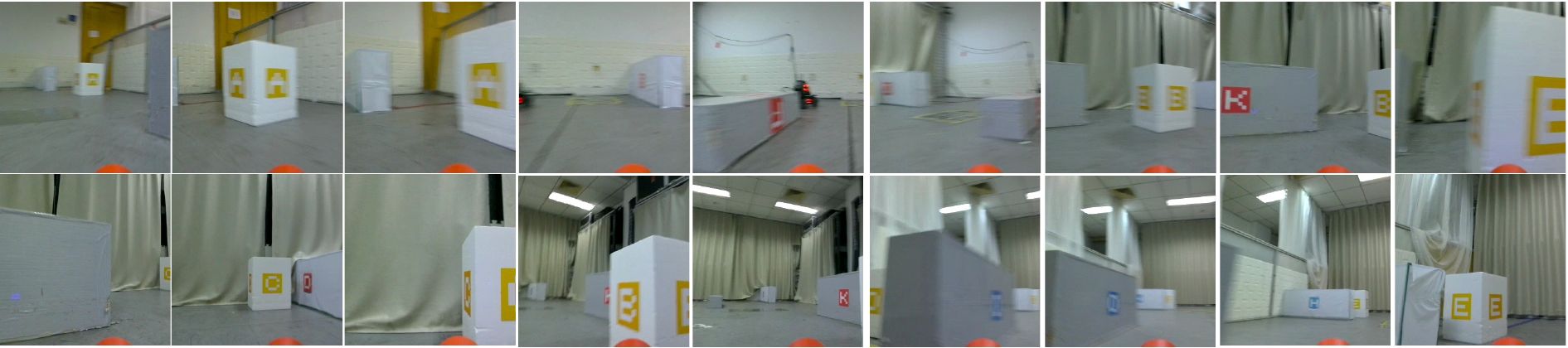}
    \caption{First-person view of navigation trajectory. We illustrate key frames and the robot successfully reaches A,B,C,D,E in order.}
    \label{FigureSim2Real}
    \vspace{-0.5cm}
\end{figure*}

\subsection{Generalization Across Scene}
In real-world applications, robots are often requested to work in new environments.
Therefore, zero-shot generalization across different scenes is an important ability for a navigation policy.
We test the performance of all the methods in an unseen environment with a different layout (as shown in Fig \ref{ExpEnvs}).
The speed limit is set to $(v_{x}=1m/s,v_{y}=1m/s, \omega=\pi / 4 \ rad/s)$ and the noise scale is set to $\delta^{shift} \sim U(-0.5,0.5) $, 
$\delta^{p}_{t} \sim U(-0.1,0.1)$, and $\delta^{p}_{t} \sim U(-0.1,0.1)$.
We report the metrics in the test scene in Table \ref{TabScene}.
Since the \textit{Pose+Laser} is simply overfitting the training environment, 
it fails in the testing environment and gets zero success rate.
The spatial information is crucial for navigation skills. 
Even though the agents are trained with only one scene, the other three methods all show generalization ability to novel scenes.
This encourages us that the egocentric map is an appropriate state space design for navigation tasks.
And our method shows 80\% success rate on the new scene.
We believe that the proposed method can achieve better generalization performance by training with more scenes and this is a direction for our future work.
By then, it can serves as a motion planning module and directly embed into the SLAM framework.

\begin{table}
    \centering
    \caption{Performance on novel scene.}
    \begin{tabular}{c|cccc}
        \toprule[1pt]
        \multicolumn{5}{c}{Performance on Test Scene} \\
        \midrule[0.5pt]
        \textbf{Methods} & Success(\%) & Activation & Collision(s) & NavTime(s)\\
        \midrule[0.5pt]
        Pose + Laser & 0.0 & - & -	& - \\
        \midrule[0.5pt]
        Map + Laser & 72.2 & 4.19 & 10.02 & \textbf{31.98}  \\
        \midrule[0.5pt]
        Teacher & 61.5 & 3.68 & 10.48 & 45.28 \\
        \midrule[0.5pt]
        Ours & \textbf{80.2} & \textbf{4.46} & \textbf{6.82} & 33.67 \\
        \bottomrule[1pt]
    \end{tabular}
    \label{TabScene}
    \vspace{-0.5cm}
\end{table}

\subsection{Sim-to-Real Generalization}
\begin{figure}[!ht]
    \centering
    \includegraphics[width=0.45\textwidth]{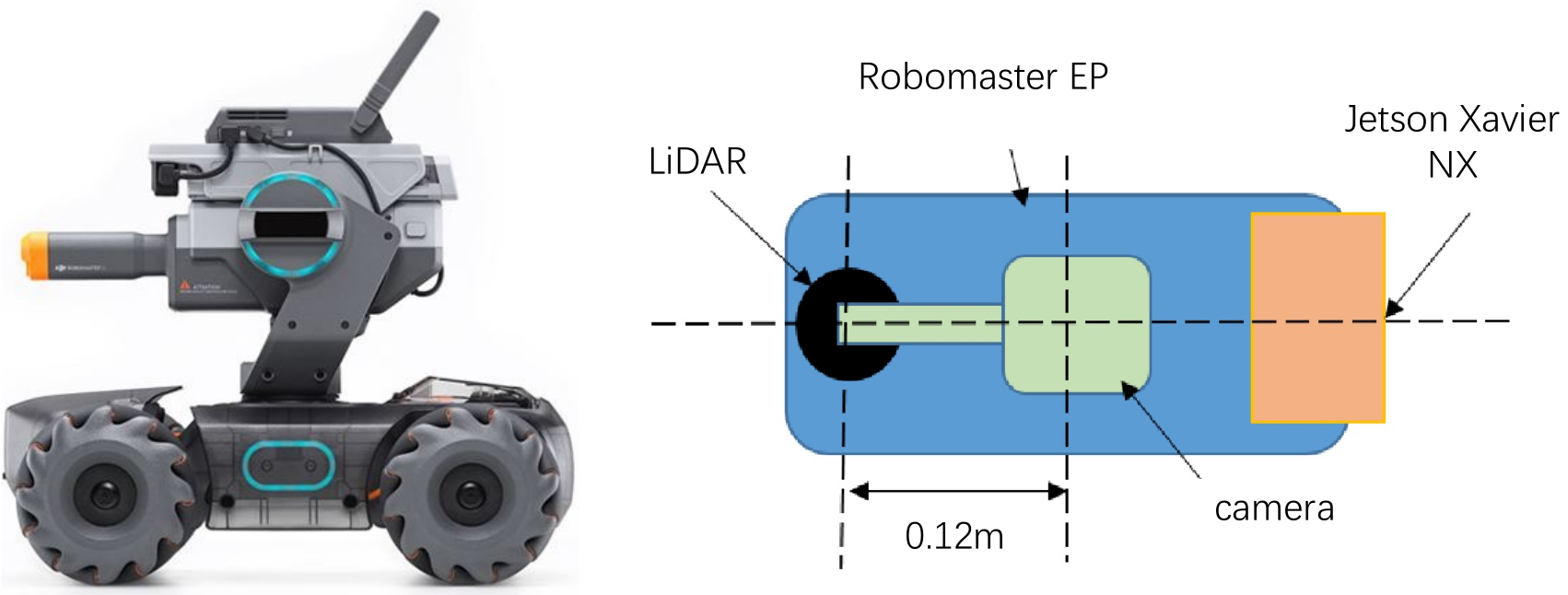}
    \caption{Overview of the Robomaster EP robot.}
    \label{FigureEP}
    \vspace{-0.5cm}
\end{figure}

The real-world experiments are implemented on a DJI RoboMaster EP robot with an RPLiDAR S2 and a nvidia Xavier NX module.
An overview of the EP robot is shown in Fig \ref{FigureEP}.
in the real world experiment, the layout is the same as the training environment.
To get the pose information,we use the point-cloud matching with an existing map to acquire the robot pose. 
Mismatch can happen, so it will lead to a different distribution of pose estimation noise.
A first-person view of the navigation trajectories are shown in Figure \ref{FigureSim2Real}. 
The robot successfully reaches A,B,C,D,E. For more information, please refer to the attached video.

\subsection{Ablation Study}
\begin{figure}
    \centering
    \includegraphics[width=0.45\textwidth]{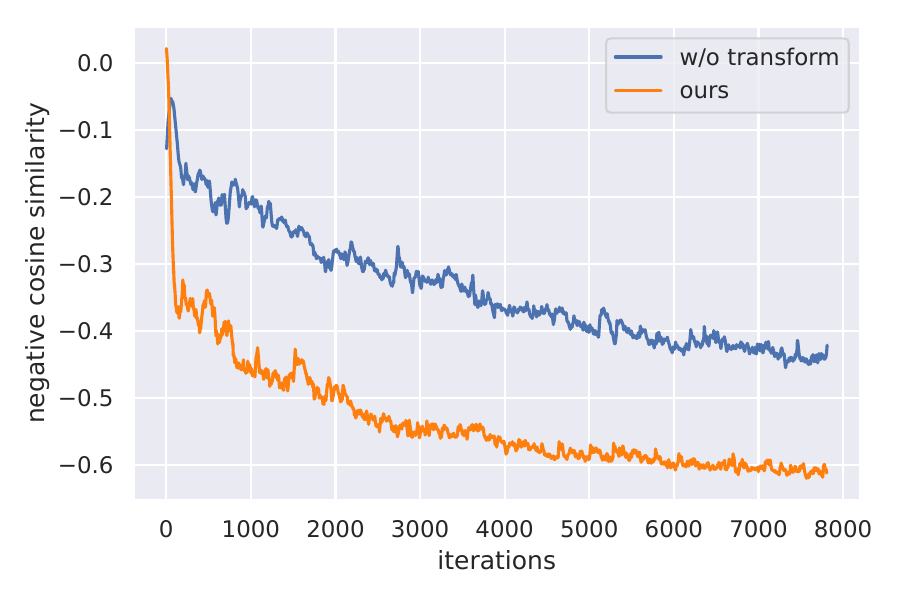}
    \vspace{-0.5cm}
    \caption{Negative Cosine Similarity in training.}
    \label{FigureCosine}
    \vspace{-0.2cm}
\end{figure}

We make three ablation studies to understand the contribution of each component.
In the first ablation, we remove the transform layers but directly use the CNN features of the egocentric map to finish the feature distillation.
We show the negative cosine similarity during the training in Fig \ref{FigureCosine}.
And we notice that without the laser signal, the similarity converges at around 0.4, but our multi-modal fusion can get more than 0.6, 
this proves that laser signal can provide additional information for feature restore and alignment.
In the second ablation, we remove the contrastive objective in teacher module.
In the third ablation, we remove the feature distillation objective in the student module.
We test the navigation performance of the above methods in the settings same as section E. 
The results are shown in Table \ref{TabAblation}.
Notably, without contrastive objective, the teacher still perfectly solves the navigation task in the ideal environment, 
but it fails to teach a good student. And the performance also decreases w/o feature distillation.
All the ablation studies suggest that selecting an proper representation space is an essential problem. 

\begin{table}
    \centering
    \caption{Ablation Study without feature distillation or without contrastive objective.}
    \begin{tabular}{c|cccc}
        \toprule[1pt]
        \multicolumn{5}{c}{$\delta^{shift} \sim U(-0.75,0.75),\delta^{p}_{t} \sim U(-0.15,0.15),\delta^{l}_{t} \sim U(-0.15,0.15) $} \\
        \midrule[0.5pt]
        \textbf{Methods} & Success(\%) & Activation & Collision(s) & NavTime(s)\\
        \midrule[0.5pt]
        w/o ft. distill & 72.5 & 4.15 & \textbf{3.32} & 36.95 \\
        \midrule[0.5pt]
        w/o cont. & 56.1 & 3.88 & 4.01 & 42.71  \\
        \midrule[0.5pt]
        Ours & \textbf{81.2} & \textbf{4.59} & 3.45 & \textbf{32.48}  \\
        \bottomrule[1pt]
    \end{tabular}
    \label{TabAblation}
    \vspace{-0.5cm}
\end{table}

\section{Conclusion}
We propose a multi-modal fusion method to incorporate the information from laser signal and map information.
By training in a teacher-student distillation framework, our proposed CMFD method is able to compensate for the pose estimation noise 
and greatly improve the generalization capability across a diverse of working conditions.
With sim-to-real experiments, we prove that our method is suitable for realistic navigation problems.
But in our work, we haven't consider the dynamic obstacles. 
How to represent the locomotion information about the obstacles and incorporate a real-time mapping module
for obstacle avoidance and navigation is one of our concerns for future works.

\clearpage
\bibliographystyle{IEEEtran}
\bibliography{root-blx}

\addtolength{\textheight}{-12cm}   




\end{document}